# Adaptive Motion Gaming AI for Health Promotion

**Pujana Paliyawan, Takahiro Kusano, Yuto Nakagawa, Tomohiro Harada, Ruck Thawonmas**

Intelligent Computer Entertainment Lab, Graduate School of Information Science and Engineering,
Ritsumeikan University, Shiga, Japan
Pujana.P@gmail.com, is0212kf@ed.ritsumei.ac.jp, is0127kh@ed.ritsumei.ac.jp,
harada@ci.ritsumei.ac.jp, ruck@is.ritsumei.ac.jp

**Abstract**

This paper presents a design of a non-player character (AI) for promoting balancedness in use of body segments when engaging in full-body motion gaming. In our experiment, we settle a battle between the proposed AI and a player by using FightingICE, a fighting game platform for AI development. A middleware called UKI is used to allow the player to control the game by using body motion instead of the keyboard and mouse. During gameplay, the proposed AI analyze health states of the player; it determines its next action by predicting how each candidate action, recommended by a Monte-Carlo tree search algorithm, will induce the player to move, and how the player' health tends to be affected. Our result demonstrates successful improvement in balancedness in use of body segments on 4 out of 5 subjects.

## Introduction

Motion gaming by using motion capture devices such as Kinect has been a center of attention for health promotion through the means of game playing. However, although motion gaming provides health benefits, we should also pay attention to its adverse effect such as repetitive strain injury and muscle imbalance that might happen when some parts of the body are too much used (Rössler et al. 2014). For development of sustaining well-being, it is also important to balance use between the right and the left sides of the body; it is noted that muscle imbalance—which leads to discomfort, injury, and some other physical ailments causing aches and pains—is commonly found in those who perform one-sided-type sports such as tennis (Maffetone 2015).

As a solution, we propose an AI that encourages the player to use the left and the right sides of the body in a balance fashion during motion gaming. The proposed AI can recognize any behavior of the player; it uses previously played gameplay data to generate the table of probability for predicting a counteraction likely to be taken by the player when it performs a certain action. Because the player executes in-game actions[1] by using body motion, AI can encourage the player to move certain segments of the body by inducing him or her to perform a desired counteraction. The proposed AI monitors accumulated movement amounts (so called "momentums" and denoted as *mm*s) on the player's body segments, and its goal is to balance the *mm*s of those segments on the left side to those on the right side of the body.

In this experiment, a fighting game called FightingICE is used for settling a battle between the player and the proposed AI. Because FightingICE does not support Kinect input, a middleware called UKI (Paliyawan and Thawonmas 2017) is used for integration of full-body control with the game. We add new mechanisms to UKI for monitoring the player's movement and assessing the state of player health during gameplay; these data are fed as input to the AI. The proposed AI searches for an action that optimally promotes the player health.

## Backgrounds

### Games for Health and Motion Gaming

Throughout the history of games research, it has been asserted that video games can be used to offer an effective and attractive means for providing exercise and rehabilitation to people of any age (Baranowski et al. 2016). Exergames is an important innovator and leading voice in healthcare. Nevertheless there are also, of course, critics and questions on whether it is really necessary and effective while we still have traditional means of exercise such as gym-based exercise and sports (Barry et al. 2016).

In order to answer the above question and understand how G4H became this successful, we will look back to problems and facts underlying today's society. It was reported that only about 20% of American adults meet Physical Activity Guidelines and less than 30% of high school students get at least 60 minutes of physical activity every day (CDC). Childhood obesity has more than doubled in children and quadrupled in adolescents in the past 30 years; one-third of children and adolescents are overweight or obese (CDC). Looking on other statistics, about a half of Americans (155 millions) play video games and about 29% of videogame players are 18 years old or younger (ESA 2015). These facts are telling us that use of video games is potential means that we can make exercises reachable to these people.

---

[1] action: an in-game attack or move, such as light punch and heavy punch. In this paper, "actions" refer to AI's actions, while the player's actions are referred to as "counteractions."

It is important to understand that G4H is not designed to be a replacement of traditional exercises or sports, but as a substitute for sedentary activities, especially those spent using a device such as a computer, television, or games console (Raymond 2013). Using many types of sources of exercises is to help ones achieve physical activity guidelines. Exergaming can invoke moderate levels of physical exercise intensity with positive feelings about exercises and reduced perception of effort (Barry et al. 2016); it is possible to used game with the objective on behavior changes, that is to say improving health habits of players as well as lilting them an interest in exercises and sports (Baranowski et al. 2016).

**UKI for full-body motion gaming**

Kinect has been recognized among gaming devices for its potential in providing full-body motion games for health promotion and rehabilitation. A study on 109 articles has reported that its possibilities and future work for rehabilitation applications are extensive (Da Gama et al. 2015). However, since motion games are usually developed specifically and their development requires more time and effort; only few games are available in the market, and a narrow variety of genres makes motion gaming reach just a niche group of gamers.

The UKI project has been launched with a goal to provide middleware that can facilitate integration of full-body control with any existing games and applications (Paliyawan and Thawonmas 2017). Besides such integration, we also add several features to enhance use, such as a module that allows the user to introduce new motions to the system by only performing them (Paliyawan et al. 2015) and a module for monitoring health risks based on an ergonomic tool REBA (Paliyawan and Thawonmas 2016).

**FightingICE for health promotion**

Properties of fighting games that enhance outcomes of exergaming are given as follows:

- **Adaptability of the difficulty level** of the exercise to fit each individual player. When the difficulty fits the player, it would optimize health outcomes, reduce frustration, and sustain interest of the player. This property can be effectively implemented in fighting games by offering a proper opponent AI for each game difficulty.

- **Clear victory condition** is an important psychological factor in exergaming. If the player identifies himself/herself as winner, it could lead to positive social comparisons and enhanced competition, serving as encouragement for continuing gameplay.

- **Simplicity** is preferred while detailed tutorials may contribute to frustration and boredom. Fighting games are simple and can be played without tutorials.

A health promotion AI is implemented for FightingICE, the fighting game platform for AI development and competition that is organized and maintained by our laboratory. Since 2014, an AI competition using FightingICE has been hold annually by the IEEE Conference on Computational Intelligence and Games (CIG), which is the premier annual event for researchers applying computational and artificial intelligence techniques to games. We believe FightingICE along with UKI can serve as a potential application for Games for Health research.

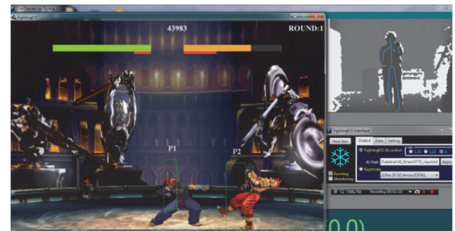

*Figure 1: UKI with FightingICE*

**Health Promotion AI**

Health Promotion AI (HP-AI) is our contribution in this paper; it is a result from integration of several concepts for health promotion from a series of our previous work with AI development. A framework for assessing players' health during full-body motion gameplay has been introduced (Paliyawan and Thawonmas 2016); we presented use of UKI for analyzing the amount of body movement and assessing postural risks on segments of body. As the first step towards what we said in that work, in the paper we focus on analysis of postural risks to balancing use of body segments, which is said to be an important key to healthy intensive exercises (Maffetone 2015); this seems perfectly suits for motion gaming using fighting games.

In this work, we present HP-AI by using MctsAi (Yoshida et al. 2016) as a base algorithm. It is noted that in many AI competitions, Monte-Carlo tree search (MCTS)-based AIs are ranked top, and the winner AI in FightingICE competition 2016 is a MCTS-based AI. MctsAi is a sample MCTS-based AI provided by our lab and is the third strongest AI among the 2016 competition entries.

**Related Work**

We have conducted a survey on existing work that uses Kinect for health promotion. Their details are summarized to Table 1. In addition, for more information please refer to Da Gama's paper (Da Gama et al. 2015). It is noted that AIs for health promotion is a new concept and to the best of our knowledge there is not yet motion game AIs, especially fighting game AIs, for health promotion.

*Table 1: Comparison of ours and existing work: **[1]** Sato et al. 2015 **[2]** Zaitsu et al. 2015 **[3]** Kayama et al. 2013 **[4]** Borghese et al. 2013 **[5]** Maloney et al. 2015 **[6]** Baranowski et al. 2011 **A** provding health benefits **B** preventing health risks / injuries during use of system **C** offering adjustable difficulty / customization.*

|   | A | B | C | Description |
|---|---|---|---|---|
| Ours | ✓ | ✓ | ✓ | promote balance use of body segments |
| [1] | ✓ | ✓ | - | improve walking, muscular strength, and balance in elderly people |
| [2] | ✓ | - | - | provide sit-to-stand exercise |
| [3] | ✓ | ✓ | - | improve balance ability and mobility, which are risk factors for falls |
| [4] | ✓ | ✓ | ✓ | provide game engine for rehabilitation |
| [5] | ✓ | - | ✓ | share lessons learned from using serious videogames in health behavior changes |
| [6] | ✓ | - | - | evaluate outcomes from playing games on children's diet, physical activity and adiposity |

## Proposed AI

System overview for controlling the proposed Health Promotion AI is shown in Figure 2. The AI analyzes the player's health state in real-time and uses supporting data from databases for determining its next move (i.e., the optimal action). Such determination gives the first priority to improvement of the player's health state, followed by strength of action. In this section, we first provide definition and details on computation of two fundamental data used in the system: (1) momentum of body movement (2) action-to-counteraction probability. We then describe how AI obtains data and processes them to determine its next action.

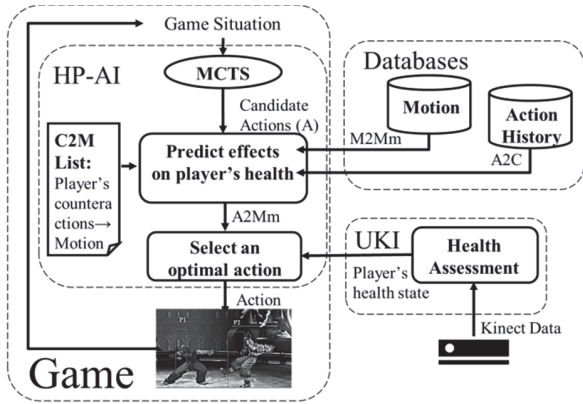

*Figure 2: System overview*

### Data and databases
**Momentum of body movement**
Raw data captured by Kinect are 3D positions ($x$, $y$, and $z$) of 20 body joints. First, coordination is localized to make data invariant to standing position of the player. Joints on upper-body are centered to center coordination of shoulder, while those on lower-body are centered to center one of hip. Second, relative change between each pair of consecutive frames is computed by using Euclidian distance. Third, joints on the center of body are omitted, and remaining joints are grouped into 4 segments that are a pair of arms and a pair of legs (Figure 3). Changes of joints in the same segment are summed up to a change of segment. Finally, changes on a segment of interest over time are accumulated and represented by a momentum in Eq. (1).

$$mm = \sum \sqrt{(x_i - x_{i-1})^2 + (y_i - y_{i-1})^2 + (z_i - z_{i-1})^2} \quad (1)$$

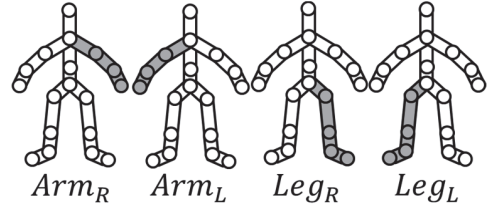

*Figure 3: Four body segments*

Motion database contains information on how much player's body segments move when he or she performs a certain motion (Motion → Momentums is denoted as M2Mm). Each record stores momentums corresponding to a motion used for executing a game skill. From example data in Figure 4, details on how motions evoke momentums are in the table M2Mm, while details on what skill (i.e., player's counteractions) is executed by using what motion are shown in the table C2M. There are 24 motions used in this experiment; for each motion, we let an experienced UKI user perform it three times to obtain sample data. As in our previous study, we found that motion data collected from one subject can be applied to others (Paliyawan et al. 2015), the proposed system does not require that every player must provide sample motion data for the sake of simplicity of use. On each file, we compute momentum as a total change, or to say we accumulate changes until the end of file. For each motion in the database, momentums are averaged from three sample data files.

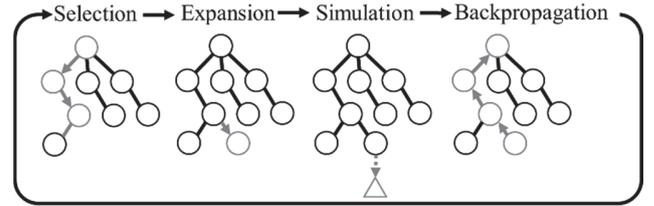

Figure 4: Data used in the system

**Action-to-counteraction probability**
Action-to-counteraction probability (A2C) is computed by using gameplay log; it represents probability on which counteraction the player tends to use when the AI takes a certain action. A counteraction is the player's first action after the AI's action. Because one action possibly leads to various counteractions, probability is used to represent a set of player's counteractions toward a given AI's action. A2Cs for all possible actions in the game are stored in a database namely Action History Database (see Figure 4)— this database is built by using log data of 45 battle rounds: 5 subjects, 9 rounds/subjects.

**Processes for determining optimal action**
In this section, we describe MCTS and processes in the system (rounded rectangles in Figure 2).

**Monte-Carlo Tree Search (MCTS)**
We embed a Monte-Carlo Tree Search (MCTS) module from MctsAi (Yoshida et al. 2016) to our AI. This module analyzes game state and recommends $n$ candidate actions to the proposed AI, in which one out of them will be selected as an optimal action. The search module provides the best action by considering strength of action under a given game situation. It is noted that if $n$ is set to 1, HP-AI will only use the strongest action, and if the value of $n$ is too large, HP-AI will be weaken; value of $n$ can be used to control game difficulty and its value is set at 3 in this study.

MCTS is a combination of tree search algorithm and Monte-Carlo method; it uses random sampling in exploration of the decision space. There are four major steps in MCTS: selection, expansion, simulation and backpropagation. The four steps are repeated until a given amount of time is elapsed. An overview of MCTS is shown in Figure 5, where the root node represents the current game situation while child nodes represent actions. A path from a root node to a leaf node is a sequence of AI actions.

Figure 5: An overview of MCTS

- **Selection**: UCB1 is employed as the selection criterion of nodes. Reward used for evaluation is computed by using changes in hit points before and after the actions is executed (denoted as $HP_{after} - HP_{before} = \Delta HP$); hitpoints considered are that of the AI and that of the player. The selection criterion is given as Eq. (2); considering the $i$-th node, $C$ is a balance parameter, $N_i$ is the number of visits at that node, $N_i^P$ is the number of visits at its parent node, and $X_i$ is the average reward (see (3) and (4)).

$$UCB1_i = \bar{X}_i + C\sqrt{\frac{2lnN_i^P}{N_i}} \qquad (2)$$

$$\bar{X}_i = \frac{1}{N_i}\sum_{j=1}^{N_i} eval_j \qquad (3)$$

$$eval_j = \Delta HP_j^{AI} - \Delta HP_j^{Player} \qquad (4)$$

- **Expansion**: By the time a leaf node is reached, if the depth of the path is shallower than a threshold and the number of visits of the leaf node is larger than a threshold, child nodes will be created from the leaf node.

- **Simulation**: A simulation is done by using a sequence of actions in the path from the root node to the leaf node as AI actions. Consequentially, it uses random actions of the same number of those in the path for the opponent's actions.
- **Backpropagation**: An update from simulation is performed to obtain UCB1 for nodes that were traversed in the path.

**Predicting effects on player's heath**

We use a searching module from MctsAi for analyzing game situation and recommending three candidate actions, in which one out of the three will be selected as optimal action of the AI. A searching module provides candidate actions by considering strength of actions under a given game state. For each candidate action (denoted as A), the AI predicts its effects on momentums of the player's body segments (denoted as A2Mm). Computation is done by using A2C, C2M, and M2Mm in respective order. In addition, the table A2Mm for all actions may also be built in advance and stored permanently during the gameplay to reduce processing time.

**Health assessment**

During gameplay, UKI accumulates momentums of the player's body segments over time from the time the game starts. A set of accumulated momentums is referred to as "Actual Momentums," or shorten as AM in Eq. (5)—for example, the first element or momentum in AM (denoted as $am_1$) is $mm_{Arm_R}$, which refers to a momentum of the right arm of the player. The proposed AI computes expected momentums (EM) for body segments by using AM, where momentum of a certain segment is expected to be equal to its pair on the opposite side of the body; for each segment, we use the maximum momentums in its pair as the expected momentums as shown in Eq. (6). Gap, denoted in Eq. (7), is then computed as a set of differences from EM to AM; from this set, it is known which segments of the body should move more and how much should it moves. These data are sent from UKI to the proposed AI as player's current health state for further processing.

$$AM = \{mm_{Arm_R}, mm_{Arm_L}, mm_{Leg_R}, mm_{Leg_L}\}$$
$$= \{am_1, am_2, am_3, am_4\} \quad (5)$$

$$EM = \{\max(am_1, am_2), \max(am_1, am_2) \\ \max(am_3, am_4), \max(am_3, am_4)\}$$
$$= \{em_1, em_2, em_3, em_4\} \quad (6)$$

$$Gap = \{gap \in \mathbb{R} \mid gap_s = em_s - am_s\} \quad (7)$$

**Select an optimal action**

The final process takes A2Mm and player's current health state as inputs. The goal of the AI is to maximize balancedness of exercise, where such balancedness (denoted as *Bal*) is computed by Eq. (8). Value of *Bal* is in the range of [0, 1], where a value of one indicates that two side of the body move in a perfect balanced fashion.

$$Bal = 1 - 2 \times \frac{\sum_{s=1}^{4} gap_s}{\sum_{s=1}^{4} em_s} \quad (8)$$

As decrease in *gaps* leads to increase in *Bal*, the AI evaluates fitness of each candidate action by estimating how much *gaps* will decrease when it performs that action (Eq. (9)). From an example in Figure 4, if Skill_A is performed by the AI, it is predicted that gaps will decrease by about 5.04, and that will improve *Bal* from 86.92% to 88.16%.

$$dec = \sum_{s=1}^{4} gap_s - \sum_{s=1}^{4} |gap_s - mm_s| \quad (9)$$

## Evaluation and Result

We evaluate effectiveness of AIs on improving balancedness of exercise. There are five subjects involve in this experiment. Each of them plays FightingICE for 9 rounds (3 matches, 3 rounds per match) against McstAI and 9 rounds against HP-AI.

By using collected data, we first analyze *Gap* between actual and expected momentums of player during gameplay. Lines in Figure 6 represent summations of actual and expected momentums on four body segments; their values are accumulated over time. *Gap* is a space between two lines. It is obvious that *Gap* is smaller when the player fights against HP-AI.

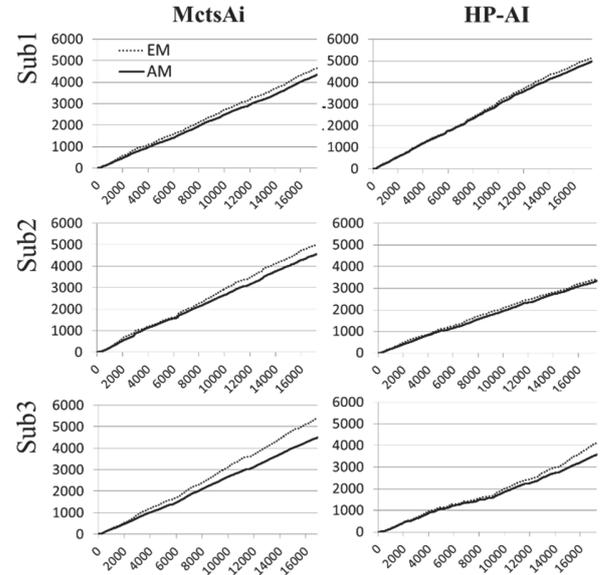

*(Continues)*

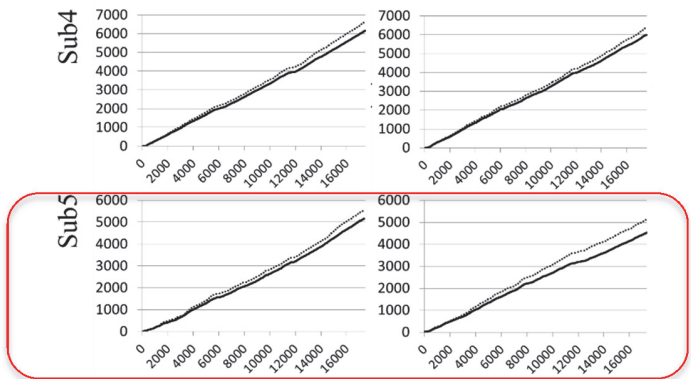

*Figure 6*: Difference between AM and EM over time (25 fps). The horizontal axis represents the number of frames that increments with the approximated rate of 25 fps. The vertical axis is value of momentum computed by Eq. (1); it represents the total amount of body movement in a unit of meter, which is summed from 4 segments or 14 joints as shown in Figure 3.

*Table 2*: Balancedness of exercise (Bal) computed at the end of gameplay.

|         | Sub1   | Sub 2  | Sub3   | Sub4   | Sub5   |
|---------|--------|--------|--------|--------|--------|
| MctsAi  | 86.51% | 82.56% | 66.30% | 85.58% | 86.17% |
| HP-AI   | 92.74% | 94.80% | 73.03% | 88.56% | 76.26% |

## Conclusion

We have presented system design and architecture for developing a game AI that promotes health of the player. Our result demonstrates possibility for the development of fighting game AI that can, in an effective manner, recognize player's behavior, analyze player's health state, and determine actions that will induce player to move in a healthy way. Our future work includes designing of additional modules for assessing and improving other health factors, such as postural risk, amount of energy expenditure, and the level of physical activity. To design health summary reports and their visualization for promoting user's motivation and leading to sustainable well-being is also a challenging topic.

## Acknowledgement

We would like to express special thanks to Makoto Ishihara for his explanations and details on how MCTS-based AIs operate.